\documentclass[5pt]{article}

\usepackage{spconf}
\usepackage{times}
\usepackage{graphicx}
\usepackage[caption=false]{subfig}
\usepackage{amsmath,amssymb}
\usepackage{algorithm,algorithmic}
\usepackage{color}
\usepackage{float}
\usepackage{subeqnarray}
\usepackage{cases}
\usepackage[american]{babel}

\usepackage[pagebackref=true,breaklinks=true,colorlinks,bookmarks=false]{hyperref}

\usepackage{spconf,amsmath,epsfig}
\usepackage{enumitem}

\usepackage{amstext}

\usepackage{fancyhdr}
\begin{document}

\title{Kill Two Birds with One Stone: Boosting Both Object Detection Accuracy and Speed with Adaptive Patch-of-Interest Composition}

\name{Shihao Zhang$^1$, Weiyao Lin$^{1*}$, Ping Lu$^2$, Weihua Li$^2$, Shuo Deng$^2$}
\address{$^1$Department of Electronic Engineering, Shanghai Jiao Tong University, China\\
$^2$Cloud Computing \& IT Institute, ZTE Corporation, China\\
($^*$Corresponding Author: wylin@sjtu.edu.cn)}

\maketitle

\begin{abstract}
  Object detection is an important yet challenging task in video understanding \& analysis, where one major challenge lies in the proper balance between two contradictive factors: detection accuracy and detection speed. In this paper, we propose a new adaptive patch-of-interest composition approach for boosting both the accuracy and speed for object detection. The proposed approach first extracts patches in a video frame which have the potential to include objects-of-interest. Then, an adaptive composition process is introduced to compose the extracted patches into an optimal number of sub-frames for object detection. With this process, we are able to maintain the resolution of the original frame during object detection (for guaranteeing the accuracy), while minimizing the number of inputs in detection (for boosting the speed). Experimental results on various datasets demon-strate the effectiveness of the proposed approach. The project page for this paper is available at \href{http://min.sjtu.edu.cn/lwydemo/Dete/demo/detection.html}{http://min.sjtu.edu.cn/lwydemo/Dete/demo/detection.html}
\end{abstract}


\section{ INTRODUCTION AND RELATED WORKS\label{section:  INTRODUCTION AND RELATED WORKS}}
 Object detection is of increasing importance in many applications including content understanding,media retrieval and intelligent transportation. In object detection, one major challenge is the tradeoff between two contradictive factors: detection accuracy and detection speed.

 Most researchers focus their researches on improving the detection accuracy. Early works try to find proper hand-crafted features in order to improve the accuracy, such as DPM ~\cite{7}, HOG ~\cite{1}and CENTRIST ~\cite{8}. The performances for these methods are oftenrestrained since hand-crafted features have limitations in effectivelycapturing the complex characteristics of objects.With the advances in deep convolutional networks (ConvNets),ConvNet-based detec-tion methods have shown big improvements on detection accuracy and have become the mainstream approaches for object detection~\cite{9}-~\cite{10},~\cite{13}. However, many ConvNet-based approaches have high computation complexity, which obviously limits their applications.

 In order to reduce the complexity of ConvNet-based detection, some speed-up methods are proposed, which improve detection speed by directly regressing object locations(e.g., SSD ~\cite{3}, YOLO~\cite{4}) or extracting object proposal regions \& features after convolution (e.g., Faster-RCNN ~\cite{5}). However, in order to guarantee the speed of convolution computation, most existing works need to perform down-sampling on the input video frames, which obviously reduces the visual information of small objects, leading to reduced detection performances. On the other hand, simple ways to maintain input frame resolutions, such as directly inputting original-resolution frames or dividing into sub-frames \& performing recognition respectively, will greatly increase the complexity of ConvNet computation, resulting in obviously reduced speed. Therefore, it is still an unsolved yet challenging problem to maintain the resolution of input information while guaranteeing the object detection speed.
\begin{figure*}[t]
  \centering
  \includegraphics[width=1\textwidth]{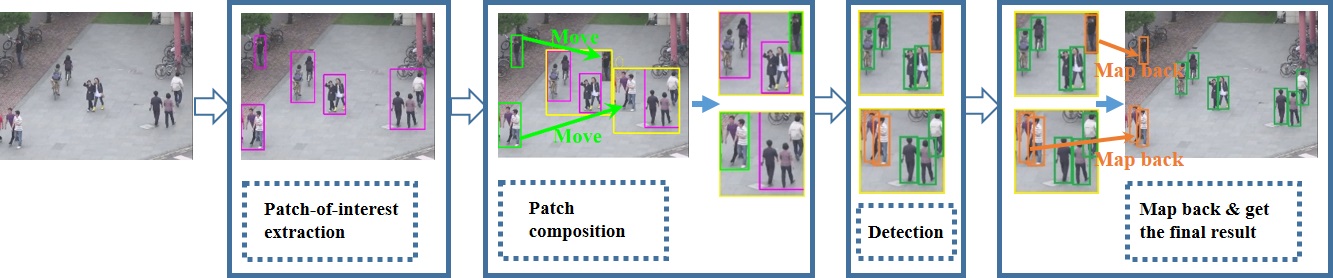}

  \caption{Framework of the proposed approach. (a) The input image. (b) Detected patches. (c) The patch composition (left) and sub-frames(right). (d) Detection results on sub-frames. (e) Map back and get the final result on the original image.}
  \label{fig:figure1}
\end{figure*}
 In this paper, we propose a new adaptive patch-of-interest composition approach for boosting both the accuracy and speed for object detection. Our approachfirst extracts patches in a video frame which have the potential to include objects-of-interest.Then an adaptive composition process is introduced to compose the extracted patches into an optimal number of sub-frames for object detection. With this approach, we are able to maintain the resolution of the original frame during object detection, while minimizing the number of inputs in detection, so as to guaranteeboth object detection accuracy and speed.

 The rest of this paper is organized as follows. Section 2 describes the framework of the proposed approach. Sections 3 to 4 describe the details of our proposed adaptive patch-of-interest composition approach, respectively. Section 5 shows the experimental results and Section 6 concludes the paper.

\section{OVERVIEW OF OUR APPROACH\label{section:OVERVIEW OF OUR APPROACH}}

 The framework of our approach is shown in Fig.~\ref{fig:figure1}. We first extract patches-of-interest in an original frame, where each patch-of-interest correspond to a region including poten-tial objects-of-interests (cf. Fig.~\ref{fig:figure1}(b)). Then a patchcompo-sition process is performed, whichautomatically finds a set of optimal locations for sub-frames and moves the extracted patches into these sub-frames (cf. Fig.~\ref{fig:figure1}(c)). Finally, the composited sub-frames are input the ConvNet-based detectors to obtain detection results (cf. Fig.~\ref{fig:figure1}(d)), and the detection results in sub-frames are simply mapped back into the original frame to achieve the final result (cf. Fig.~\ref{fig:figure1}(e)).

 In our framework, patch-of-interest extraction and patchcomposition are the key components for our approach. Their details are described in Sections 3 and 4, respectively.

\section{PATCH-OF-INTEREST EXTRACTION\label{section:PATCHES-OF-INTEREST EXTRACTION}}
\begin{figure}[H]
  \centering
  \includegraphics[width=0.475\textwidth]{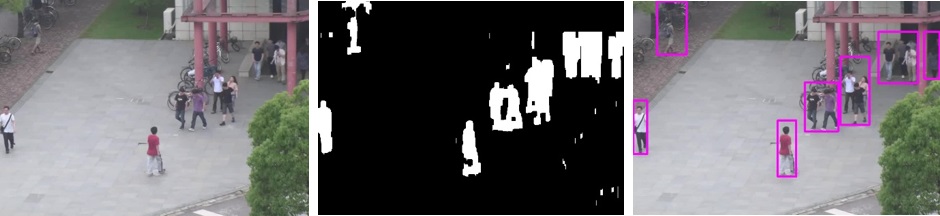}

  \caption{Procedure of extracting patches: (a) The original image, (b) The foreground after morphological filtering, (c) The image including patches.}
  \label{fig:figure2}
\end{figure}

 The patch-of-interest extraction component includes two steps: potential region detection and patch extraction. They
 are described in the following:

 \textbf{Potential region detection.} Potential region detection step aims to detect potential regions that may include objects of interest. In this paper, since we mainly focus on surveillance scenarios whose backgrounds are normally static, we use foreground extraction followed by simple morphological filtering~\cite{11},~\cite{12} to detect potential regions, as shown in Fig.~\ref{fig:figure2}. It should be noted that foreground extraction is just one way to obtain potential regions. In practice, we can also use other methods to get potential regions in various scenarios, e.g., first detect region proposals ~\cite{6} and then filter the results by a simple classifier ~\cite{2}.

 \textbf{Patch extraction.} Patch extraction step aims to identify rectangular-shaped patches that include the detected potential regions. In this paper, we simply derive a bounding box for each connected potential region as the extracted patch as shown in Fig.~\ref{fig:figure2} (c).
\begin{figure}[H]
  \centering
  \subfloat[]{\includegraphics[width=4cm,height=2.8cm]{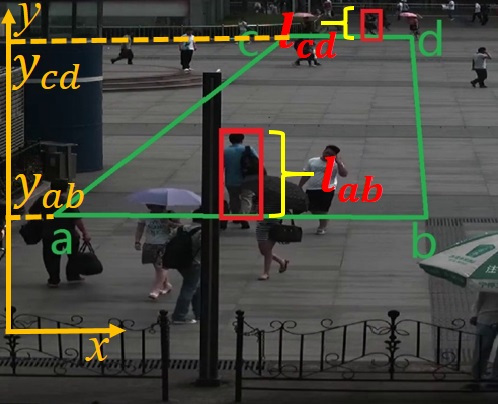}
    \label{fig:figure3a}}
  \hspace{2mm}
  \subfloat[]{\includegraphics[width=4cm,height=2.8cm]{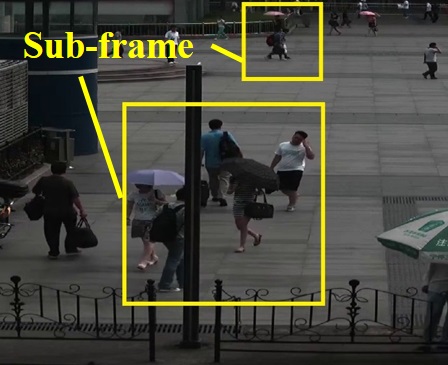}
    \label{fig:figure3b}}
  \caption{ (a) Scaling factor calculation procedure. (b) Sub-frames with different sizes.
  }\label{fig:figure3}
\end{figure}
 Two things need to be mentioned about the patch extraction step: (1) we leave a blank region of 3-pixel width on the edge of each patch, so as to guarantee reliable detection performances when the patch is composited with other patches. (2) More importantly, since object size sin a scene often vary a lot due to their different distances to a camera, we give patches different scaling factors according to their locations in a scene. In this way, we are able to composite patches with similar object sizes into sub-frames(cf. Section 4) and reduce the impact of large size variance in the detection process.

 The scaling factor of a patch is calculated as shown in Fig.~\ref{fig:figure3a}. Specifically, we first find a region which corresponds to a rectangle in the real scene (cf. the green rectangle in Fig.~\ref{fig:figure3a}), and measure the vertical axis $y_{ab}$ and $y_{cd}$ of the rectangle's near-end and far-end lines $\overline{ab}$ and $\overline{cd}$ in the frame.

 Then, we select two people appearing at the near and far ends of the rectangle, and measure their heights $l_{ab}$ and $l_{cd}$ in the frame (cf. the red rectangles in Fig.~\ref{fig:figure3a}). Finally, the scaling factor of an object located at vertical axis $y_{input}$ is calculated by Eq.~\ref{equation 1}:

\begin{equation}
\label{equation 1}
 \beta_{input}=k\cdot(\frac{l_{ab}-l_{cd}}{y_{ab}-y_{cd}}\cdot y_{input}+\frac{y_{ab}\cdot l_{cd}-y_{cd}\cdot l_{ab}}{y_{ab}-y_{cd}})\,
\end{equation}

 where $\beta_{input}$ is the scaling factor for object vertically located at $y_{input}$ , $k$ is a constant. In this paper, considering the ConvNet-based detector has a certain ability to detect objects of different sizes, we do not calculate $\beta_{input}$ for each object. Instead, when the object sizes vary widely, we divide an input scene into 2-3 vertical regions and use a fixed scaling factor for each region.

\section{ADAPTIVE PATCH COMPOSITION\label{section: ADAPTIVE PATCH COMPOSITION}}
 After extracting patches-of-interest, we apply an adaptive composition process to composite the extracted patches into an optimal number of sub-frames for object detection. Note that this component is the keypart of our approach.
\subsection{Objective function\label{section:Objective function}}
 Given a set of patches-of-interest extracted from a frame:$\Theta=\{P_1,P_2,...,P_{N_P}\}$, where $N_P$ is the number of patches, we aim to composite them into an optimal set of sub-frames  such that: (1) these sub-frames can include all patches (to make the detector cover all potential regions), and (2) the number of sub-frames are minimized (to reduce detection complexity). The objective function is described by Eq.~\ref{equation 2}.
\begin{small}
\begin{equation}
\begin{aligned}
\begin{split}
\label{equation 2}
 \Omega_{F}^{*}=arg\min_{N_F}\max_{F_1..F_{N_F}}
 \frac{\alpha\Psi(\Omega_F,\Theta_P)+\sum\limits_{j=1}^{N_F}\Phi(F_j,\Theta_P)}{H(N_F)}\,
\end{split}
\end{aligned}
\end{equation}
\end{small}
 $s$.$t$.$\forall P_i$,$\exists F_j$,$P_i$ is included by $F_j$,$i\in[1,N_P]$,$j\in[1,N_F]$

 where $\Omega_{F}^{*}=\{F_{1}^{*},F_{2}^{*},...,F_{N_{F}}^{*}\}$ is the optimal set of subframes. $\Psi(\Omega_F,\Theta_P)$ is the term measuring the suitability of sub-frame locations. $\Phi(F_j,\Theta_P)$ is the term measuring the suitability of patch distributions in a sub-frame $F_j$. $H(N_F)$ is the optimality evaluation on the number of sub-frames $N_F$.The terms $\Psi(\Omega_F,\Theta_P)$, $\Phi(F_j,\Theta_P)$, and $H(N_F)$ are detailed in the following.
\begin{figure}[H]
  \centering
  \subfloat[]{\includegraphics[width=2.8cm,height=2.6cm]{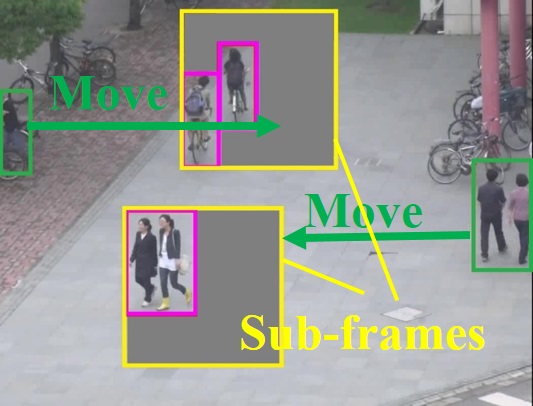}
    \label{fig:figure4a}}
  \subfloat[]{\includegraphics[width=2.8cm,height=2.6cm]{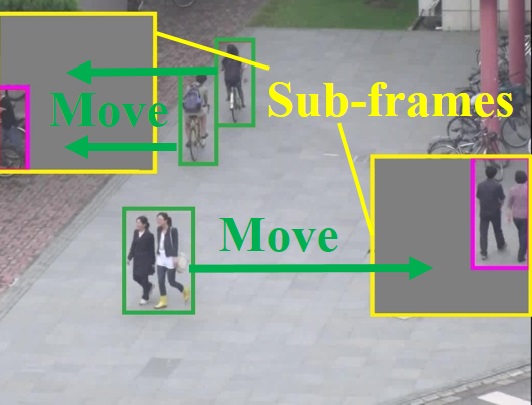}
    \label{fig:figure4b}}
  \subfloat[]{\includegraphics[width=2.8cm,height=2.6cm]{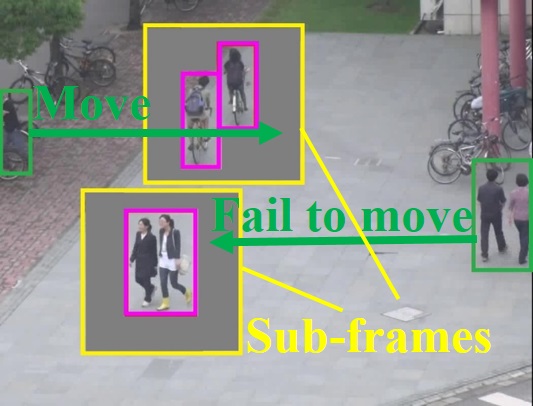}
    \label{fig:figure4c}}
  \caption{  Different locations and patch distributions of sub-frames in an image: (a) Detected patch locations with the sub-frame location term and the patch distribution term. (b) Detected patch locations with the patch distribution term, but without the sub-frame location term. (c)Detected patch locations with the sub-frame location term, but without the patch distribution term.
  }\label{fig:figure4}
\end{figure}

 \textbf{Sub-frame location term.} When compositing sub-frames, we first want to determine proper locations of sub-frames and move patches that are not covered by sub-frames into the blank regions of sub-frames (cf. Fig.~\ref{fig:figure4}). In our approach, we view locations consisting of a large number of patches-of interest with large sizes as the proper locations of sub-frames, since it can greatly reduce the number and total size of uncovered patches. Therefore, we define the term of measuring the sub-frame location as:
\begin{small}
\begin{equation}
\begin{aligned}
\begin{split}
\label{equation 3}
 &\Psi(\Omega_F,\Theta_P)=\\
 &\ln\Bigg(\big(\frac{\sum\limits_{i=1}^{N_P}h_{i}^{P}\cdot w_{i}^{P}\cdot \beta_{i}^{P}}{\sum\limits_{i=1}^{N_P}h_{i}^{P}\cdot w_{i}^{P}\cdot \beta_{i}^{P}\cdot\min(\sum\limits_{j=1}^{N_F}g(i,j),1)}-1\big)^{-1}+e\Bigg)\,
\end{split}
\end{aligned}
\end{equation}
\begin{equation}
\label{equation 4}
g(i,j)=\left\{\begin{matrix}
0,& {F}_{j}\text{ covers }{P}_{i}\\
1,& {F}_{j}\text{ not covers } {P}_{i}
\end{matrix}\right.
\end{equation}
\end{small}
 where $P_i$ represents the $i-th$ patch-of-interest, and $i\in[1,N_P],j\in[1,N_F]$. $(x_{i}^{P},y_{i}^{P})$ represents the location of $P_i$. $(w_{i}^{P},h_{i}^{P})$represents the width and height of $P_i$. $\beta_{i}^{P}$ is the scaling factor of $P_i$. $F_j$ represents the $j-th$ sub-frame. $e$ is the base of the natural logarithmic function.The numerator of Eq.~\ref{equation 3} represents the area sum of all patches in a frame, and the denominator represents the area sum of patches that are covered by any sub-frame. With Eq.~\ref{equation 3}, we are able to find suitable sub-frame locations which consist of large numbers of patches-of-interest with large sizes (cf. Fig.~\ref{fig:figure3a} and Fig.~\ref{fig:figure3b}).

 Moreover, note that since we introduce a scaling factor $\beta$ for patches at different vertical locations (cf. Fig.~\ref{fig:figure3a}, the size of sub-frames at different locations are also controlled by the same scaling factor. For example, in Fig.~\ref{fig:figure3b}, the sub-frame on the top has smaller size while the sub-frame in the bottom has larger size.

 \textbf{Patch distribution term.} The sub-frame location term in Eq.~\ref{equation 3}cannot perfectly determine the location of a sub-frame since multiple locations in a neighborhood may create the same value in Eq.~\ref{equation 3} but have different patch distributions. For example, in Fig. 4, since the sub-frames in (a) and (c) cover the same patches, they have the same value in Eq.~\ref{equation 3}. However, their patch distributions are different where patches in (a) are located closer to the border of sub-frames. Obviously, the sub-frame locations in (a) is better than (c),since sub-frames in (a) have more blank regions where more uncovered patches can be moved in, while an uncovered patch fails to be moved into sub-frames in (c).

 Therefore, we propose a patch distribution term to encourage covered patches to stay close to the border of sub-frames:
\begin{small}
\begin{equation}\label{equation 5}
\begin{aligned}
\begin{split}
&\Phi(F_j,\Theta_P)=\\
&\frac{\sum\limits_{i=1}^{N_P}\sqrt{|(x_{i}^{P}-x_{j}^{F})\cdot(y_{i}^{P}-y_{j}^{F})|}\cdot\sqrt{h_{i}^{P}\cdot w_{i}^{P}}\cdot g(i,j)}{\sum\limits_{i=1}^{N_P}g(i,j)}
\end{split}
\end{aligned}
\end{equation}
\end{small}
 where $g(i,j)$ is calculated according to Eq.~\ref{equation 4},  $(x_{j}^{F},y_{j}^{F})$is the locationofsub-frame $F_j$, $(x_{i}^{P},y_{i}^{P})$is the location of patch $P_i$, $(w_{i}^{P},h_{i}^{P})$is the width and height of $P_i$.

 \textbf{Sub-frame number term.} One major target of sub-frame composition is to find minimum number of sub-frames to cover all patches-of-interest, so as to minimize the compu-tation complexity of ConvNet-based detection. Therefore, we also define a sub-frame number term by:
\begin{small}
\begin{equation}\label{equation 6}
H(N_F)=k\cdot N_F+b
\end{equation}
\end{small}
where $k$ and $b$ are constants.
\subsection{ Optimization of the objective function}

 Since the objective function of Eq.~\ref{equation 2} is non-convex with non-linear constraints, it is difficult to directly solve Eq.~\ref{equation 2}. Therefore, in this paper, we develop an iterative optimization approachto approximately solve Eq.~\ref{equation 2}, which is able to find ideal solution with low complexity.
\subsubsection{Simplified objective function}
 Since the constraint in the original objective function in Eq.~\ref{equation 2} is complex, we utilize a simple inequality to approximate it and convert this inequality to a penalty function. The simplified objective function is described by:
\begin{small}
\begin{equation}\label{equation 7}
\begin{aligned}
\begin{split}
\Omega_{F}^{*}=arg\min_{N_F}\max_{F_1..F_{N_F}}&\frac{\alpha\Psi(\Omega_F,\Theta_P)+\sum\limits_{j=1}^{N_F}\Phi(F_j,\Theta_P)}{H(N_F)}\\ &-\delta G(\Omega_F,\Theta_P)
\end{split}
\end{aligned}
\end{equation}

\begin{equation}\label{equation 8}
\begin{aligned}
\begin{split}
&G(\Omega_{F},\Theta_{P})=\\
&\left\{\begin{matrix}
0, & \frac{\sum_{N_{F}}^{j}w_{F}^{j}\cdot h_{F}^{j}\cdot \beta _{F}^{j}}{\sum_{N_{P}}^{i}w_{P}^{i}\cdot h_{P}^{i}\cdot \beta _{p}^{i}}\geq 1, & i\in[1,N_{P}], & j\in[1,N_{F}]\\
1, & \frac{\sum_{N_{F}}^{j}w_{F}^{j}\cdot h_{F}^{j}\cdot \beta _{F}^{j}}{\sum_{N_{P}}^{i}w_{P}^{i}\cdot h_{P}^{i}\cdot \beta _{p}^{i}}< 1, & i\in[1,N_{P}], & j\in[1,N_{F}]
\end{matrix}\right.
\end{split}
\end{aligned}
\end{equation}
\end{small}

 where $\delta$ is a positive constant with a large value. $G(\Omega_F,\Theta_P)$ in Eq.~\ref{equation 8} is the inequality condition to approximate the constraint in Eq.~\ref{equation 2}. According to Eq.~\ref{equation 8}, when the area sum of sub-frames is less than that of patches, the candidate sub-frame solution $\Omega_F$ is considered as unsatisfactory. Otherwise, $\Omega_F$ is probable to hold all patches. Therefore, by optimizing Eq.~\ref{equation 7}, we are able to find a satisfactory set of sub-frames $\Omega_{F}^{*}$ whichcomprehensively consider all important factors including sub-frame location $\Psi(\Omega_F,\Theta_P)$, sub-frame number $H(N_F)$, sub-frame coverage $G(\Omega_F,\Theta_P)$, and patch distribution $\Phi(F_j,\Theta_P)$.
\subsubsection{Solving simplified objective function}

 The objective function in Eq.~\ref{equation 7} can be solved by different ways. In this paper, we develop a generic-based process ~\cite{15}to solve Eq.~\ref{equation 7}. The process includes five steps as described in the following.
\begin{algorithm}[t]
   \caption{Process of solving objective function}
   \small{
     {\bf Input}:  A set of patches $\Theta_P$ from an image
     \\{\bf Output}:   A set of sub-frames $\Omega_F$ that including all patches}\\
     \begin{algorithmic}[1]
     \small{
      \STATE Down-sample the image and determine $L_{min}$\&$L_{max}$ to initialize sub-frame locations \& number, which is calculated by Eq. (9),then generate a set of probable initial sub-frame sets $\{\Omega_F\}$.
      \STATE Update sub-frame sets $\{\Omega_F\}$  by elite retention, selection, crossing and mutation~\cite{15}.
      \STATE Update sub-frame sets $\{\Omega_F\}$ by the local search.
      \STATE Calculate the objective cost value in Eq.~\ref{equation 7}, and determine whether the iterative updating process can be terminated. If false, go back to step 2.
      \STATE Verify whether the final solution $\Omega_{F}^{*}$ in step 4 satisfies the strict constraint in Eq.~\ref{equation 2}. If true, the process ends. Otherwise, update $L_{min}$ and $L_{max}$, then return to step 1.
     }
     \end{algorithmic}
     \label{algorithm: Process of solving objective function}
\end{algorithm}

\begin{figure}[H]
  \centering\includegraphics[width=5cm,height=3.5cm]{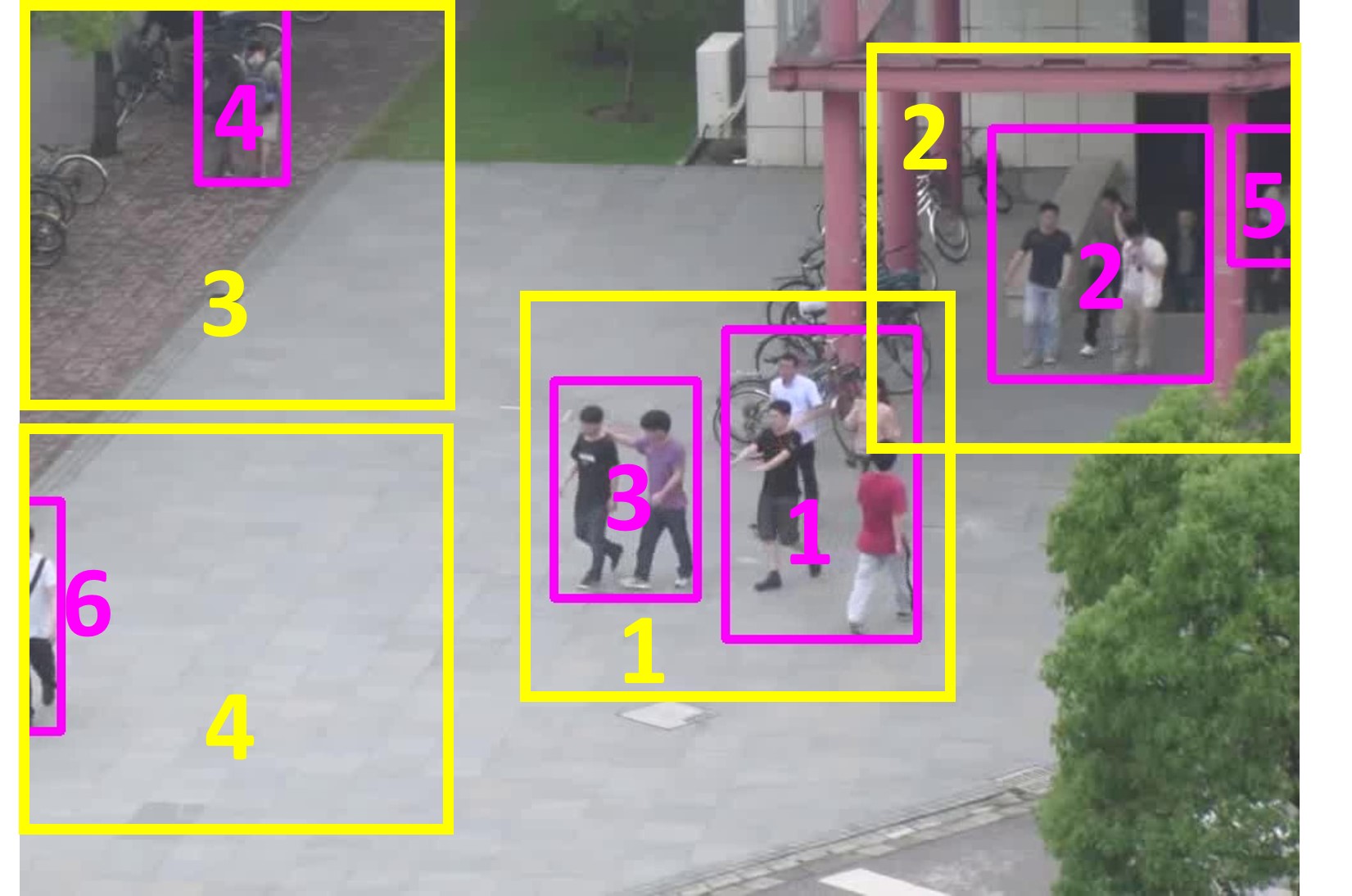}
  \caption{ Greedy strategy that adds sub-frames to cover patches-of-interest and determine the possible range of sub-frame numbers. }
   \label{fig:figure5}
\end{figure}

 \textbf{Step 1: Initializing sub-frame locations \& number.} Initializing sub-frames with proper locations and number is important to quickly find the solution of the objective function. In this paper, we apply a real number coding strategy~\cite{16} to perform sub-frame initialization, which simultaneously creates a large number of initial sub-frame sets covering the variations of sub-frame number and sub-frame locations. However, since the possible variations of sub-frame number and locations are huge, directly creating initial sets is computationally intensive. Therefore, in this paper, we introduce a sampling strategy to reduce the number of initial sets. Specifically, we first down-sample the original frame, so that the possible variation of sub-frame locations are reduced.Then, we further utilize a greedy strategy to reduce the possible value range of sub-frame numbers, as shown in Fig.~\ref{fig:figure5}.

 According to Fig.~\ref{fig:figure5}, we first sort patches-of-interest in a frame from large sizes to small sizes (cf. the red numbers in Fig.~\ref{fig:figure5}). Then, we add sub-frames to sequentially cover patches from large sizes to small ones until all the patches are covered (cf. the yellow rectangles and yellow numbers in Fig. ~\ref{fig:figure5}). Note that during the process of adding sub-frames, if an uncovered patch can be covered by any existing sub-frame, we will not add new sub-frames to cover this patch. Finally, we can determine the upper bound $L_{max}$ and lower bound $L_{min}$ of sub-frame number range through the sub-frame adding process. Specifically, when in a certain step, the total size of sub-frames exceeds the total size of patches, $L_{min}$ will be set as this sub-frame number. Similarly, $L_{max}$ is set by the sub-frame number when the total sub-frame size exceeds twice of the total patch size.

 After determining the possible range of sub-frame numbers, we can create a reduced number of initial sets to cover the variations of sub-frame number and locations, where $N_g$ is calculated by:
\begin{equation}\label{equation 9}
N_g=\alpha_3(L_{max}^{2}-(L_{min}-1)^{2})
\end{equation}
 where $\alpha_3$ is a constant, and $L_{max}$ and $L_{min}$ are the upper and lower bounds of sub-frame numbers. Compared with directly deriving initial sets from the entire range of sub-frame number \& locations, the number of initial sets in Eq.~\ref{equation 9} is greatly reduced. According to our experiments, this reduced initial set number can still properly cover the proper variation of sub-frames and create satisfactory results.

 \textbf{Step 2: Updating sub-frame sets by elite retention,sele-ction, crossing and mutation.} After obtaining initial sets of sub-frame locations and numbers, we follow similar steps as the generic process which simultaneously update all sub-frame sets through elite retention, selection, crossing and mutation operations~\cite{15} and gradually search for better results. Note that in order to prevent the best sub-frame set in one iteration from being destroyed in the next iteration, we utilize the elite retention strategy by mandatorily adding the best sub-frame set in the next iteration. Besides, in order to avoid the inclusion of too many noisy updates, we only receive the mutation result where the new sub-frame covers at least one patch.

 \textbf{Step 3:Local search updating.}Since the update process in step 2 is random, the convergence speed by step 2 is slow. In order to speed up the convergence process, we propose an additional local search strategy. Specifically, during each iteration, we let sub-frames in each sub-frame set to search in a neighborhood region and evaluate the cost value according to Eq.~\ref{equation 7}. If a better location is found, a sub-frame will be moved to this location.

 \textbf{Step 4: Termination evaluation.} After each iteration, we will check the result to see whether the iterative updating process can be terminated. Specifically, after each iteration, we calculate the objective cost value in Eq.~\ref{equation 7} for all sub-frame sets and record the best one. If the best sub-frame set does not change in four iterations, we will terminate the iteration process and use this best sub-frame sets as the optimal solution.
 Otherwise, go back to step 2.

 \textbf{Step 5: Result verification.}Since the condition of the objective function in Eq.~\ref{equation 8} is an approximation of the strict condition in Eq.~\ref{equation 2}, the optimized solution after step 4 may not perfectly satisfy the condition in Eq.~\ref{equation 2} (i.e., the derived sub-frames may not be able to completely include all patches). Therefore, we further introduce a verification process to verify whether the final solution $\Omega_{F}^{*}$ in step 4 satisfies the strict constraint in Eq.~\ref{equation 2}. Specifically, suppose the final solution $\Omega_{F}^{*}$ contains $N_{F}^{*}$ sub-frames, the verification process includes four sub-steps.
\begin{equation}\label{equation 10}
 R_k=\{R_{k1},R_{k2},...,R_{kN_R}\},k\in[1,N_{F}^{*}]
\end{equation}

 $\bullet$ Find up to $N_R$ of the largest blank rectangles in each sub-frames, which are represented as Eq.~\ref{equation 10}.

 $\bullet$ Find the uncovered patches and sort them from large sizes to small ones.

 $\bullet$ Sequentially pick out uncovered patches from large sizes to small ones, and determine whether there exists a rectangle blank region $R_{ki}$ that can contain the current patch. If not, the verification process is failed. We will go back to step 1, increase the lower and upper bounds of sub-frame numbers $L_{min}$and $L_{max}$ by 1, and find a new set of sub-frame solutions.

 $\bullet$ If all uncovered patches can be covered by the blank regions of sub-frames, the verification process is successful, and the entire optimization process is finished.

  Note that since the objective function in Eq.~\ref{equation 7} properly approximates the original objective function in Eq.~\ref{equation 2}, most solutions from step 4 can successfully pass the verification process without having to re-solve the entire optimization process. According to our experiments, the entire optimization process only takes less than $3$ms for a frame (cf. Section 5), which is computationally very efficient. The entire optimization process is summarized by Algorithm ~\ref{algorithm: Process of solving objective function}.

\section{EXPERIMENTAL RESULTS\label{section: EXPERIMENTAL RESULTS}}

\subsection{Experiments setting}

 We perform experiments on two real-scene surveillance video sequences: CANTEEN and STATION. The resolution of both sequences are $1280\cdot720$, and the number of frames in these sequences are 1212 and 1533, respectively. Some example frames for these sequences are shown in Fig.~\ref{fig:figure7} and Fig. ~\ref{fig:figure8}. Note that these sequences are challenging in that: (1) Objects (i.e., pedestrians)in both scenes are crowded and difficult to differentiate; (2) The size of pedestrians varies a lot with both large-size pedestrians and small-size ones.

 Moreover, in order to further demonstrate the effectiveness of our approach on multiple-camera scenarios, we also perform experiments on a public BEST data set ~\cite{}. Specifically, we select 4 video sequences related to the same building from BEST and sequentially stitch their frames into super frames for later detection, as in Fig.~\ref{fig:figure9}.

 We perform experiments on a PC with 15G memory, 4 GHz CPU, and a NVidia TITAN X GPU. The Single Shot MultiBox Detector (SSD) with input size $300\cdot300$ ~\cite{3}is used as the ConvNet-based detector in our framework since it has relatively high detection speed. Note that our framework is general and in practice, other detectors ~\cite{4}-~\cite{5} can also be integrated into our approach.
\subsection{Performance comparisons}

 In order to evaluate the effectiveness of our approach, we compare the following four methods.

 $(1)$ Directly down-sample the original frames into $300\cdot300$ and input into ConvNet-based detector (DS).

 $(2)$ Divide each frame into $300\cdot300$ non-overlapping sub-frames and input them into ConvNet-based detector respectively (DIV).

 $(3)$ Our approach which uses sub-frames with $300\cdot300$ sizes to cover patches-of-interest (Our-S).

 $(4)$ Our approach which uses sub-frames with $500\cdot500$ sizes to cover patches-of-interest (Our-L) and then down-samples them to $300\cdot300$ for detection. This method can be viewed as a fast version of our approach, which utilizes larger sub-frame sizes to cover more patches, so as to reduce the number of sub-frames in later detection steps.
\begin{table}[H]
\setlength{\belowcaptionskip}{10pt}
\centering
\caption{speed on the video sequence of CANTEEN}\label{tab:Table1}
\footnotesize{
\label{table1}
\begin{tabular}{|p{2cm}|c|c|c|c|}
\hline
& 1-precision& Recall& F1& Speed (frame/s)\\
\hline
DS & 0.36& 0.60& 0.62& 32.1\\
DIV & 0.25& 0.65& 0.68& 3.8\\
Our-L & 0.21& 0.64& 0.71& {\bf25.5}\\
Our-S & {\bf0.19}& {\bf0.66}& {\bf0.73}& 14.3\\
\hline
\end{tabular}}
\end{table}

\begin{table}[H]
\setlength{\belowcaptionskip}{10pt}
\centering
\caption{Results on the video sequence of STATION}\label{tab:Table2}
\footnotesize{
\label{table2}
\begin{tabular}{|p{2cm}|c|c|c|c|}
\hline				
& 1-precision& Recall& F1& Speed (frame/s)\\
\hline													
DS & 0.46& 0.36& 0.43& 29.6\\
DIV & 0.42& 0.47& 0.52& 3.7\\
Our-L & 0.41& 0.44& 0.50& {\bf23.8}\\
Our-S & {\bf0.33}& {\bf0.48}& {\bf0.56}& 13.7\\
\hline
\end{tabular}}
\end{table}

\begin{table}[H]
\setlength{\belowcaptionskip}{10pt}
\centering
\caption{Results on the BESTDATASET}\label{tab:Table3}
\footnotesize{
\label{table3}
\begin{tabular}{|p{2cm}|c|c|c|c|}
\hline				
& 1-precision& Recall& F1& Speed (frame/s)\\
\hline									
DS & 0.39& 0.35& 0.44& 28.8\\
DIV & 0.27& {\bf0.44}& 0.55& 3.3	\\
Our-L & 0.25 & 0.43	& 0.54	& {\bf25.7} \\
Our-S & {\bf0.18}& 0.42& {\bf0.57}& 24.2\\
\hline
\end{tabular}}
\end{table}

\begin{table}[H]
\setlength{\belowcaptionskip}{10pt}
\centering
\caption{Time consuming in each part of our method (ms/per frame)}\label{tab:Table4}
\footnotesize{
\label{table4}
\begin{tabular}{|c|c|c|c|c|}
\hline				
& Patch extraction& Patch composition& detection& total\\
\hline								
Our-S & 4.65& 2.62& 31.9& 39.2\\
\hline
\end{tabular}}
\end{table}

\begin{figure}[H]
  \centering
  \subfloat[]{\includegraphics[width=2.8cm,height=2.6cm]{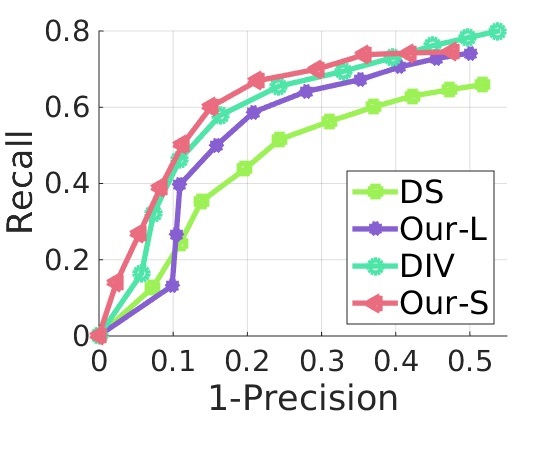}
    \label{fig:figure6a}}
  \subfloat[]{\includegraphics[width=2.8cm,height=2.6cm]{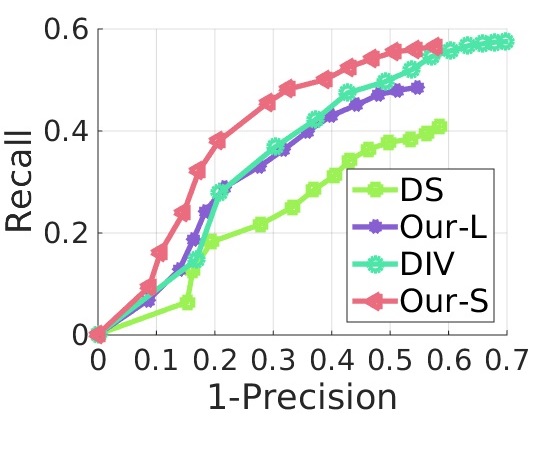}
    \label{fig:figure6b}}
  \subfloat[]{\includegraphics[width=2.8cm,height=2.6cm]{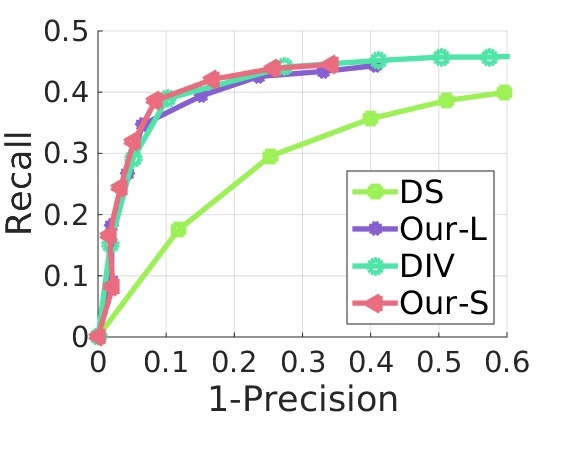}
    \label{fig:figure6c}}
  \caption{Recall vs 1-Precision Curve: (a) CANTEEN sequence; (b) STATION sequence; (c) BESTdataset.
  }\label{fig:figure6}
\end{figure}

\begin{figure}[H]
  \centering
  \subfloat[]{\includegraphics[width=2.8cm,height=2.6cm]{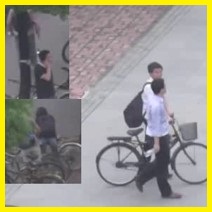}
    \label{fig:figure10a}}
  \subfloat[]{\includegraphics[width=2.8cm,height=2.6cm]{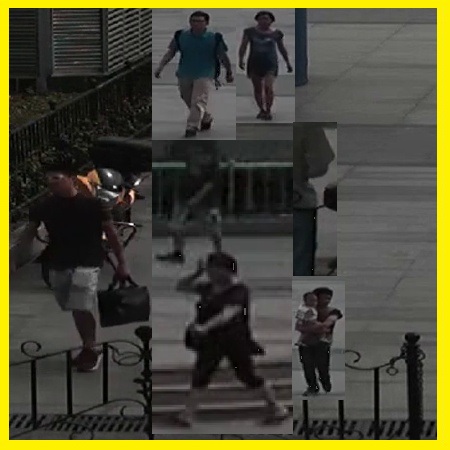}
    \label{fig:figure10b}}
  \subfloat[]{\includegraphics[width=2.8cm,height=2.6cm]{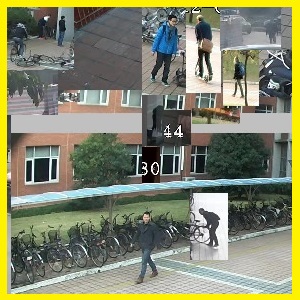}
    \label{fig:figure10c}}
  \caption{ Examples of composited sub-frames by our approach in different datasets. }
   \label{fig:figure10}
\end{figure}

\begin{figure}[H]
  \centering
  \subfloat[]{\includegraphics[width=4cm,height=2.8cm]{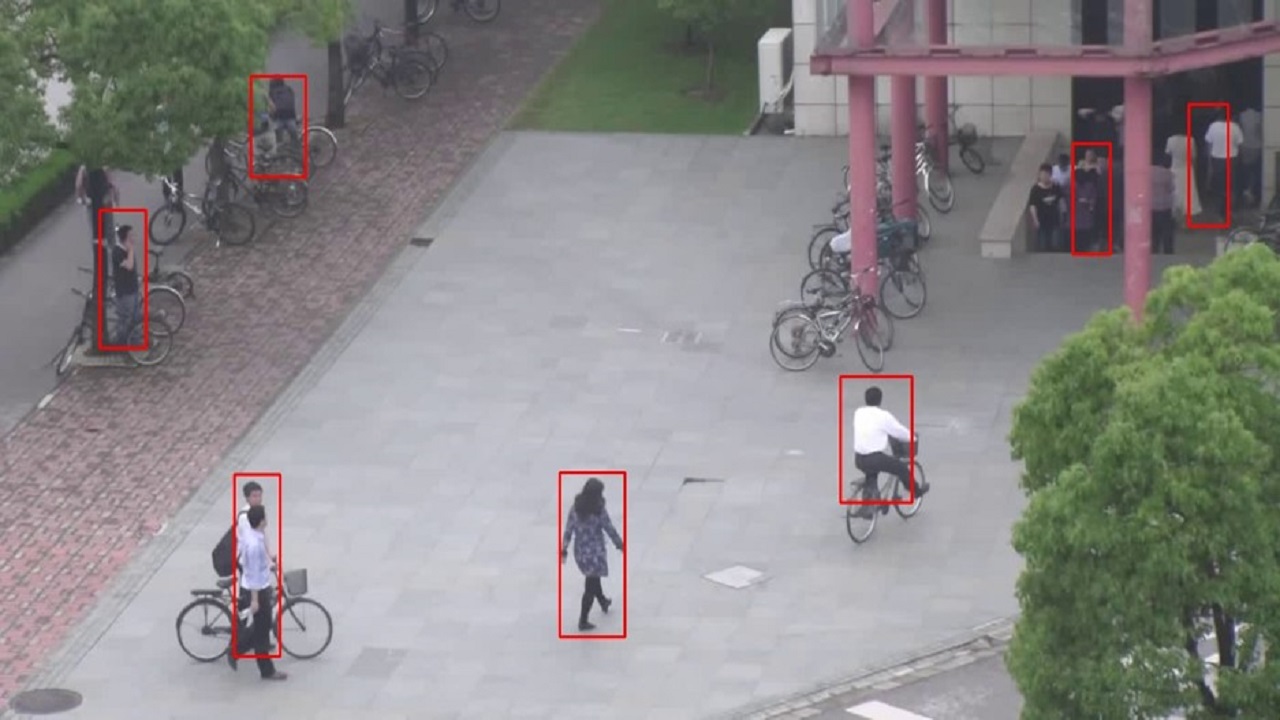}
    \label{fig:figure7a}}
  \subfloat[]{\includegraphics[width=4cm,height=2.8cm]{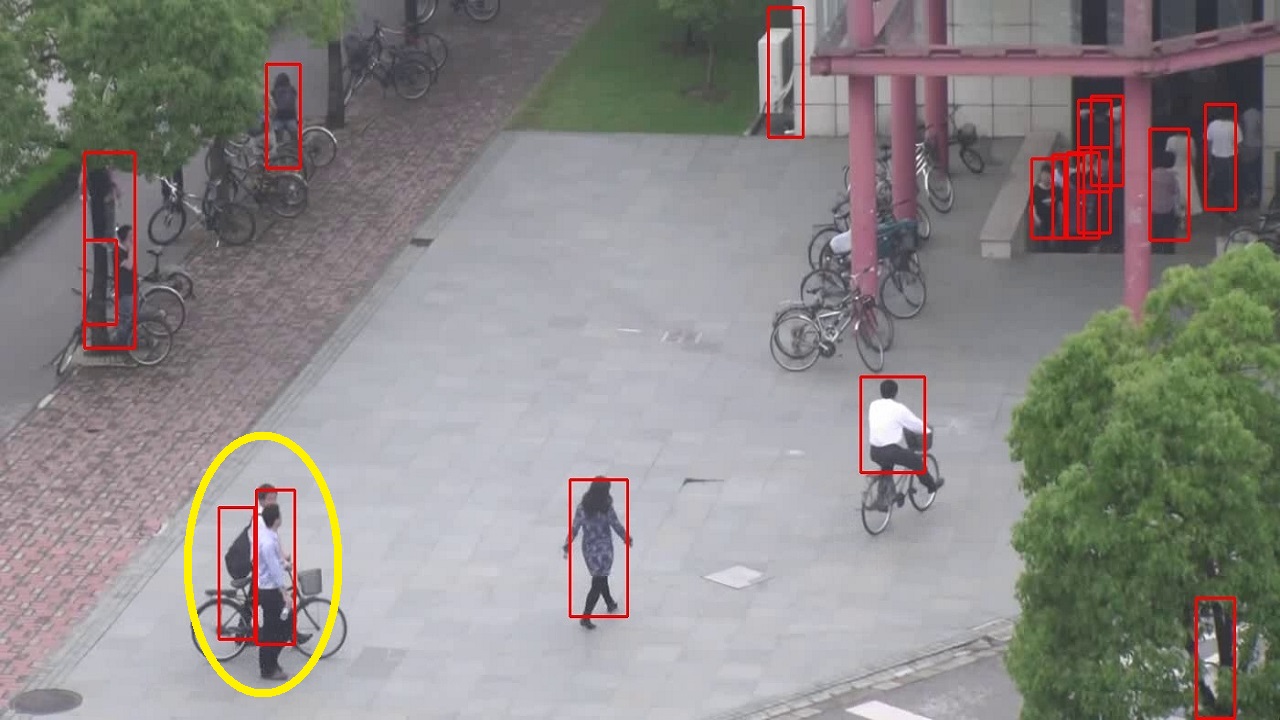}
    \label{fig:figure7b}}\\
  \subfloat[]{\includegraphics[width=4cm,height=2.8cm]{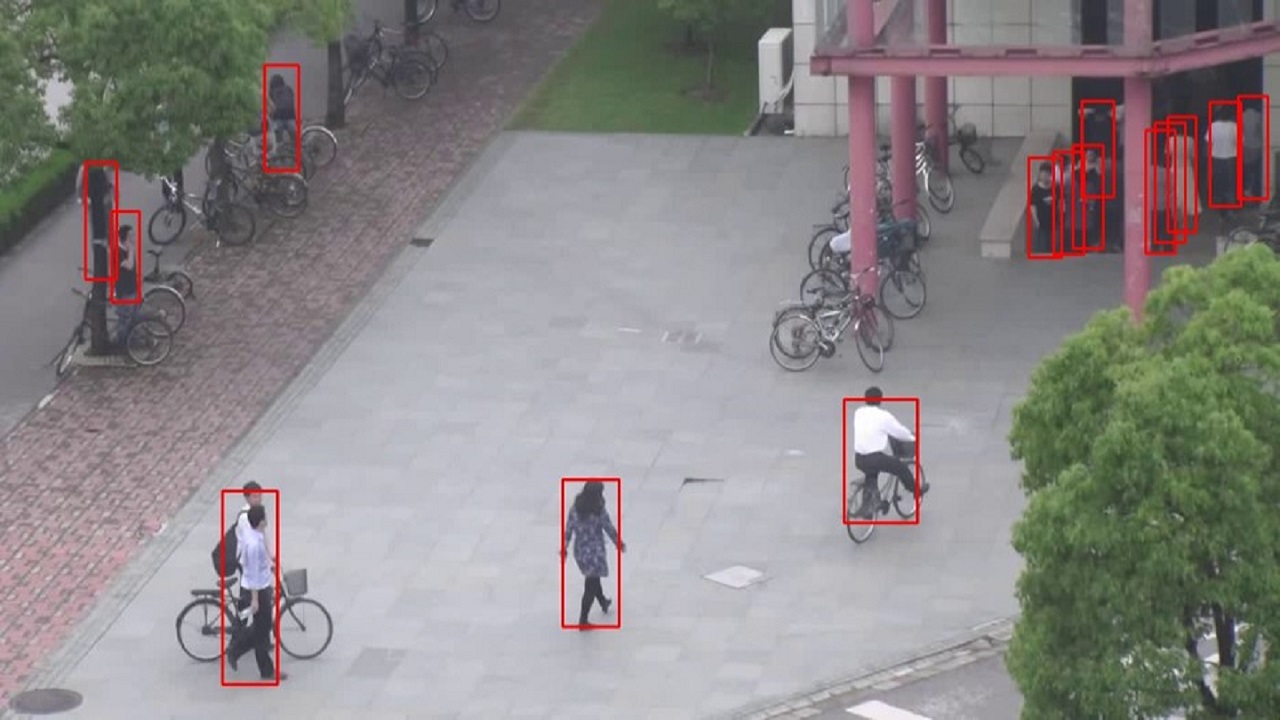}
    \label{fig:figure7c}}
  \subfloat[]{\includegraphics[width=4cm,height=2.8cm]{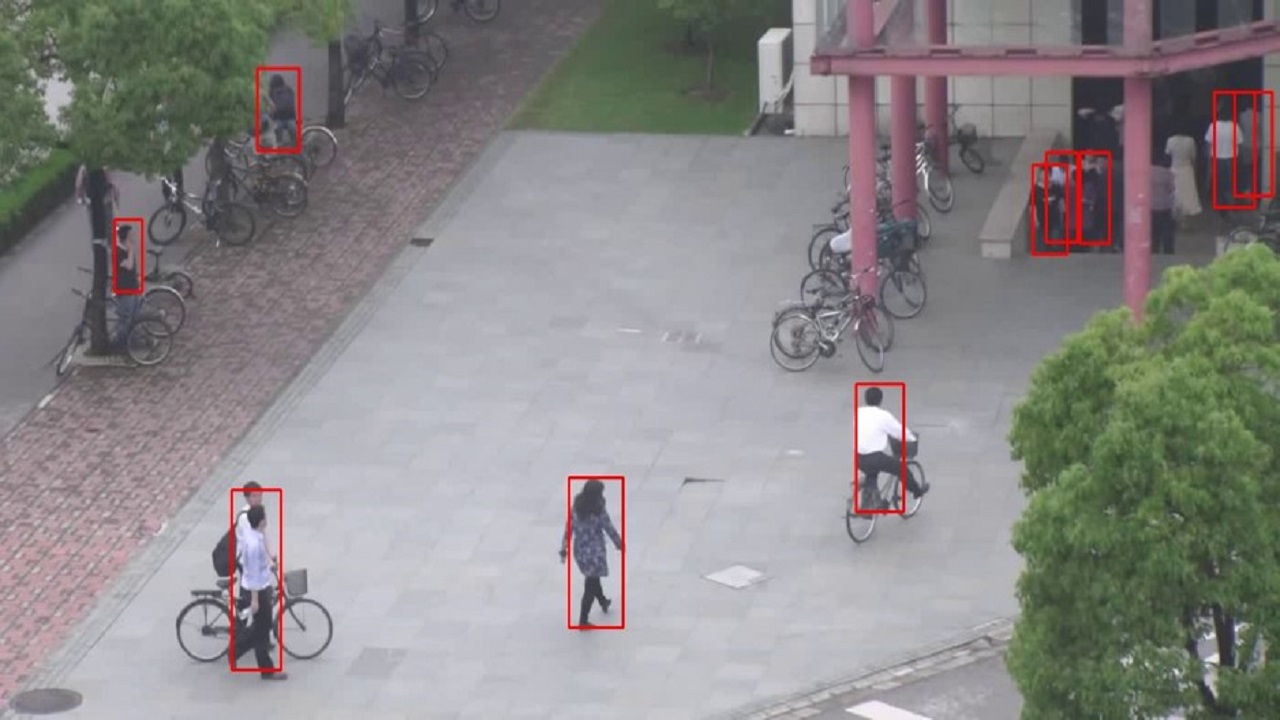}
    \label{fig:figure7d}}
  \caption{ Examples of detection results on CANTEEN sequence.
  }\label{fig:figure7}
\end{figure}

\begin{figure}[H]
  \centering
  \subfloat[]{\includegraphics[width=4cm,height=2.8cm]{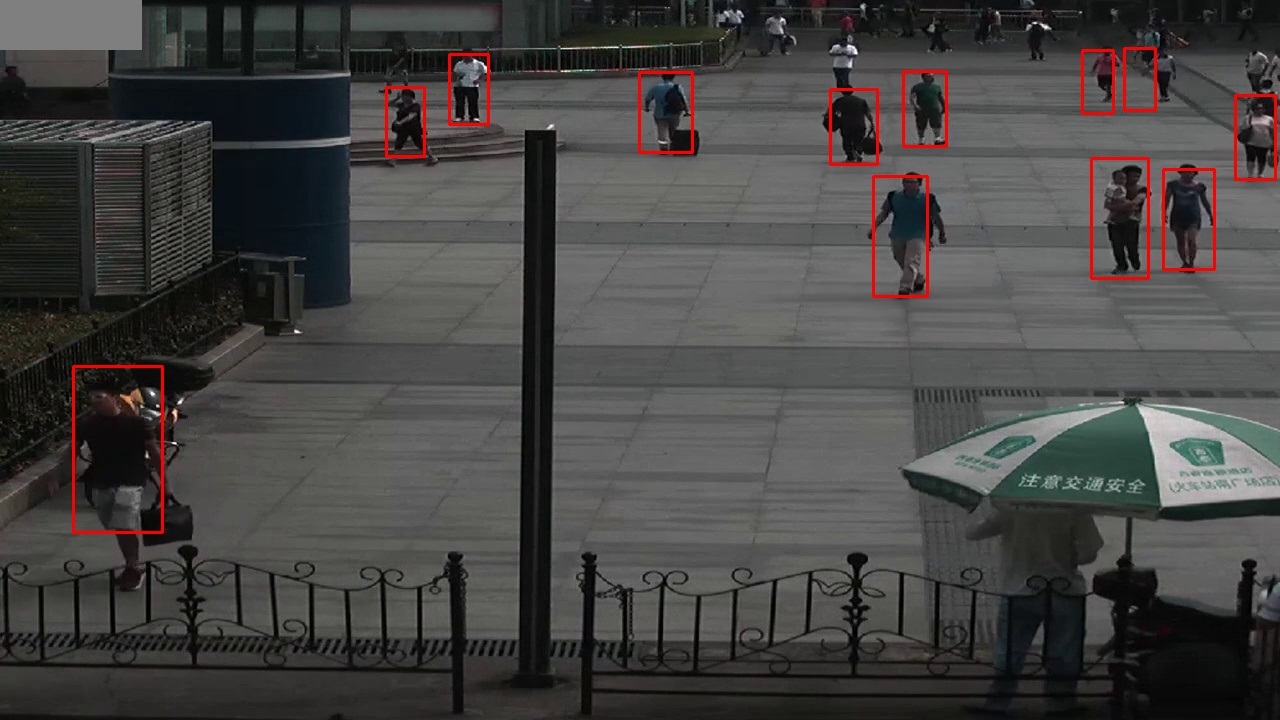}
    \label{fig:figure8a}}
  \subfloat[]{\includegraphics[width=4cm,height=2.8cm]{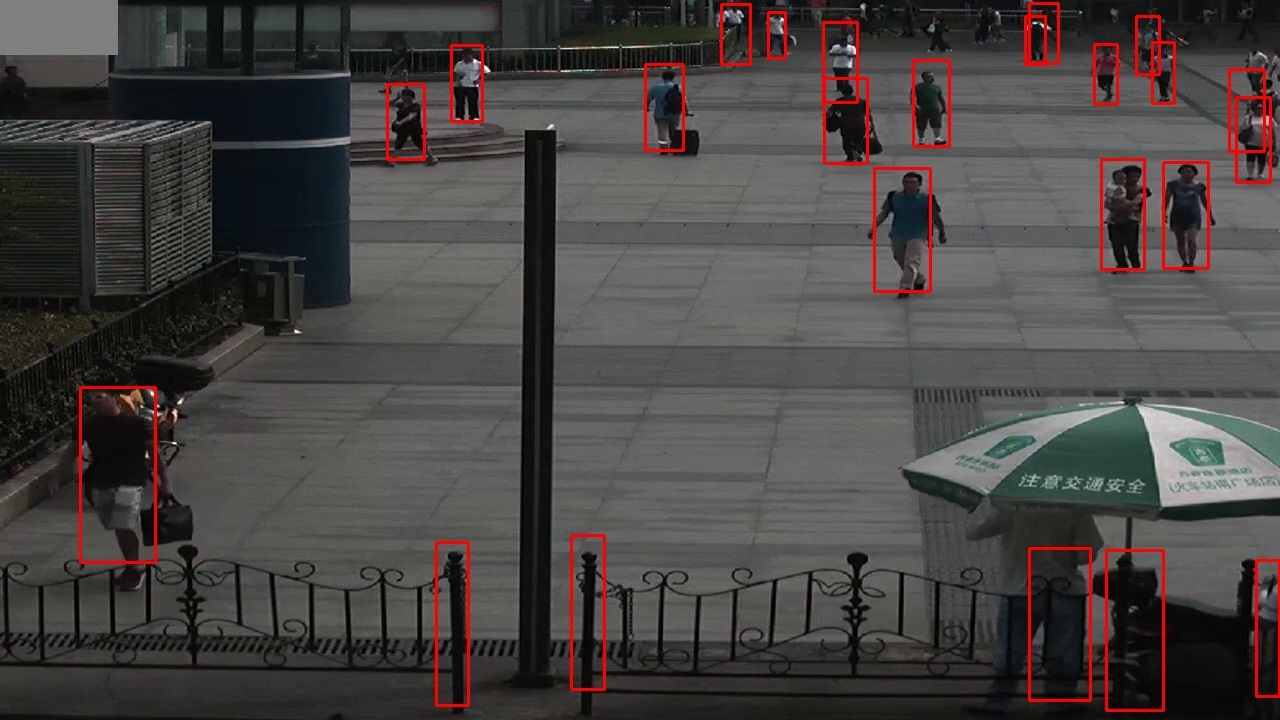}
    \label{fig:figure8b}}\\
  \subfloat[]{\includegraphics[width=4cm,height=2.8cm]{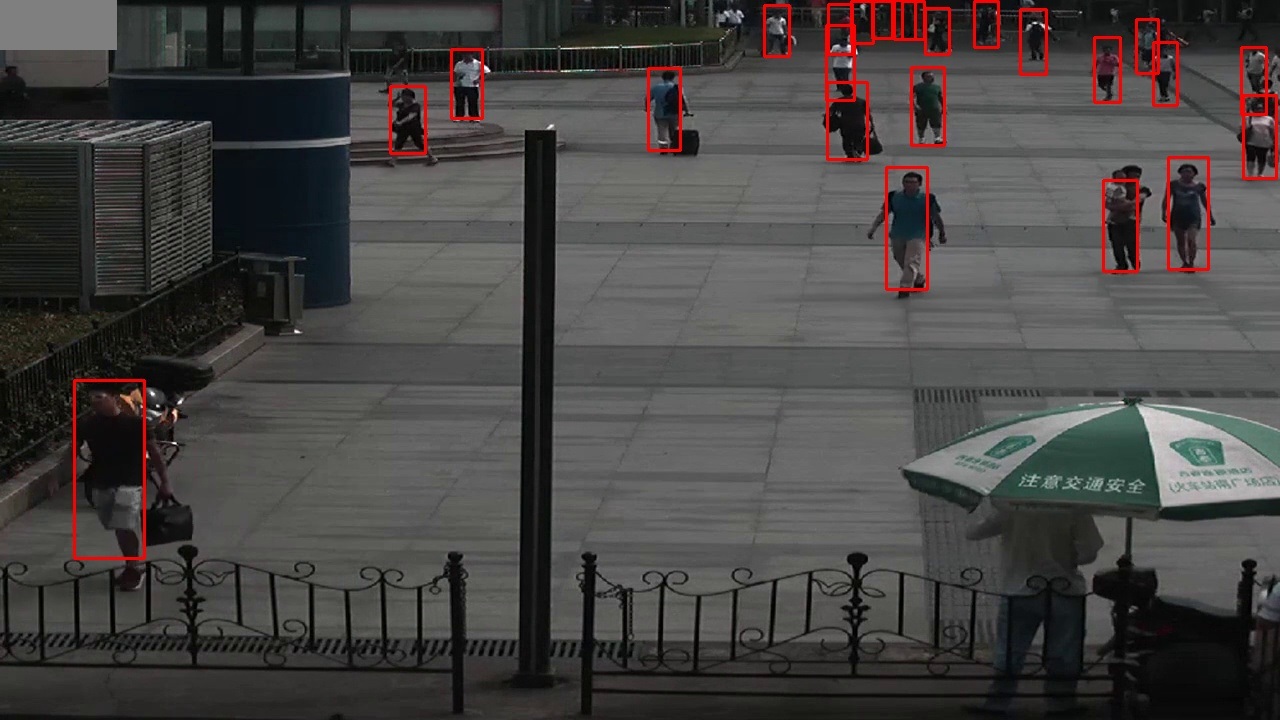}
    \label{fig:figure8c}}
  \subfloat[]{\includegraphics[width=4cm,height=2.8cm]{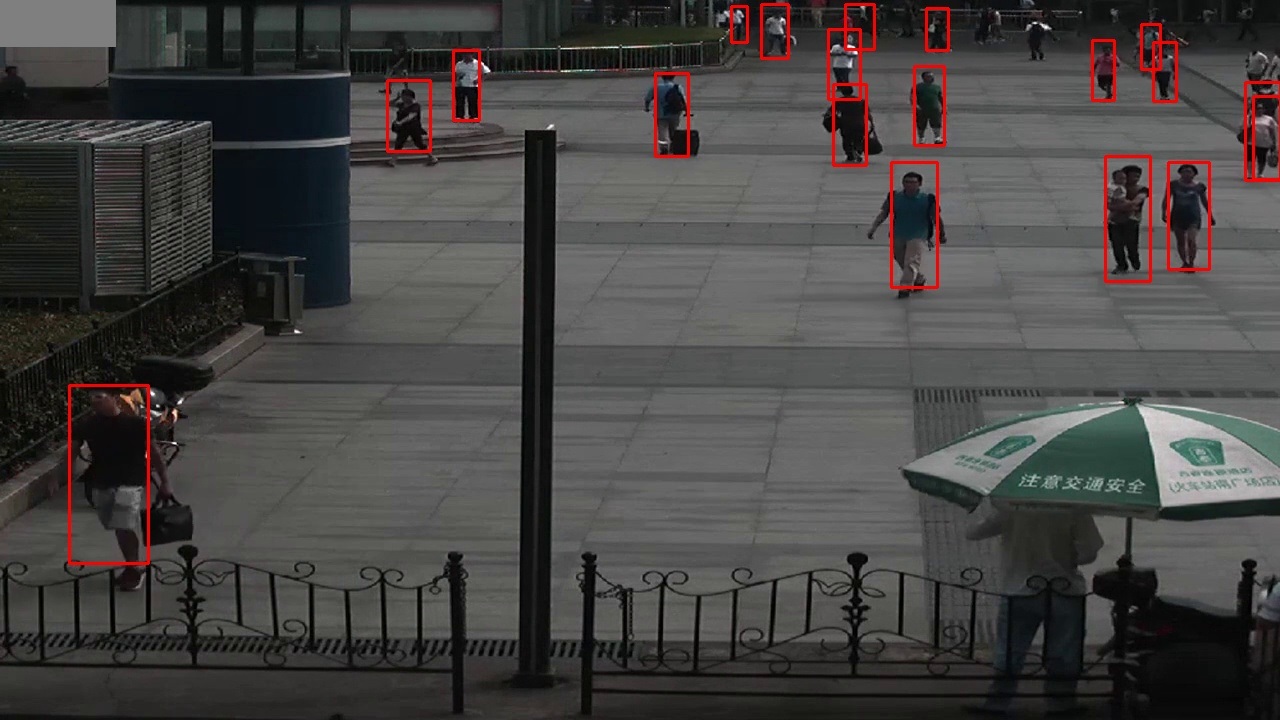}
    \label{fig:figure8d}}
  \caption{  Examples of detection results on STATION sequence.
  }\label{fig:figure8}
\end{figure}

\begin{figure}[H]
  \centering
  \subfloat[]{\includegraphics[width=4cm,height=2.8cm]{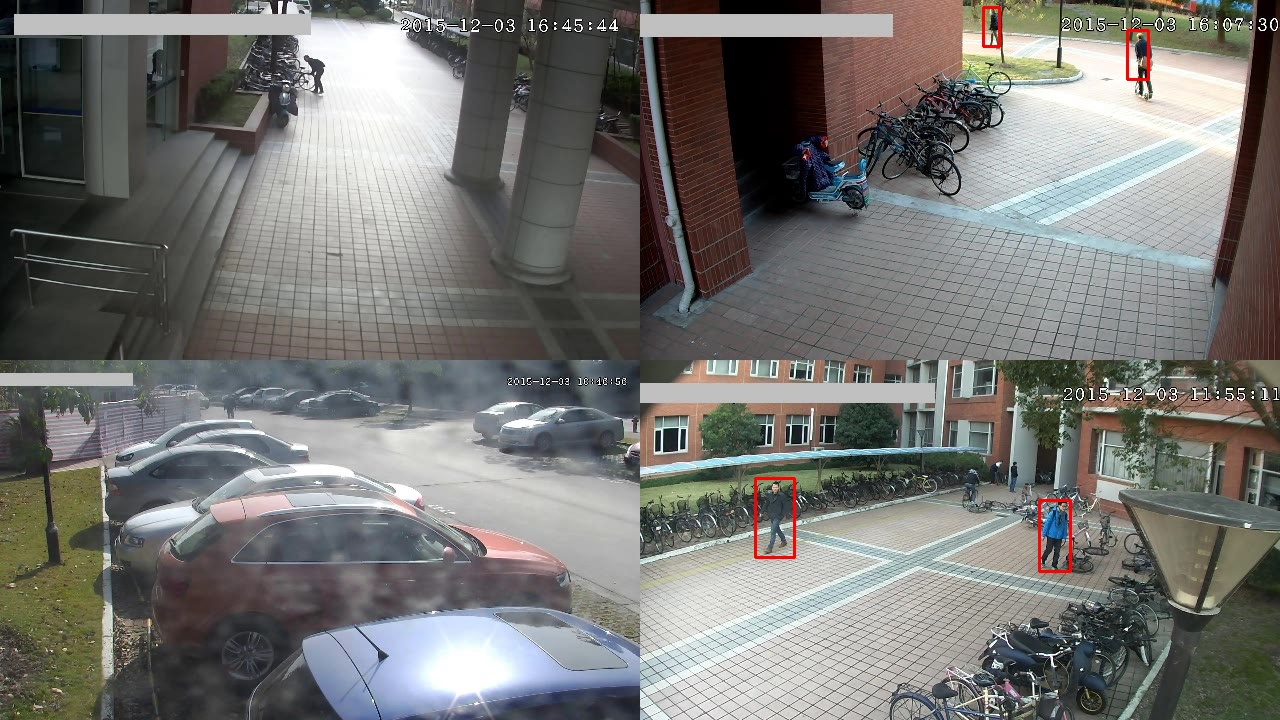}
    \label{fig:figure9a}}
  \subfloat[]{\includegraphics[width=4cm,height=2.8cm]{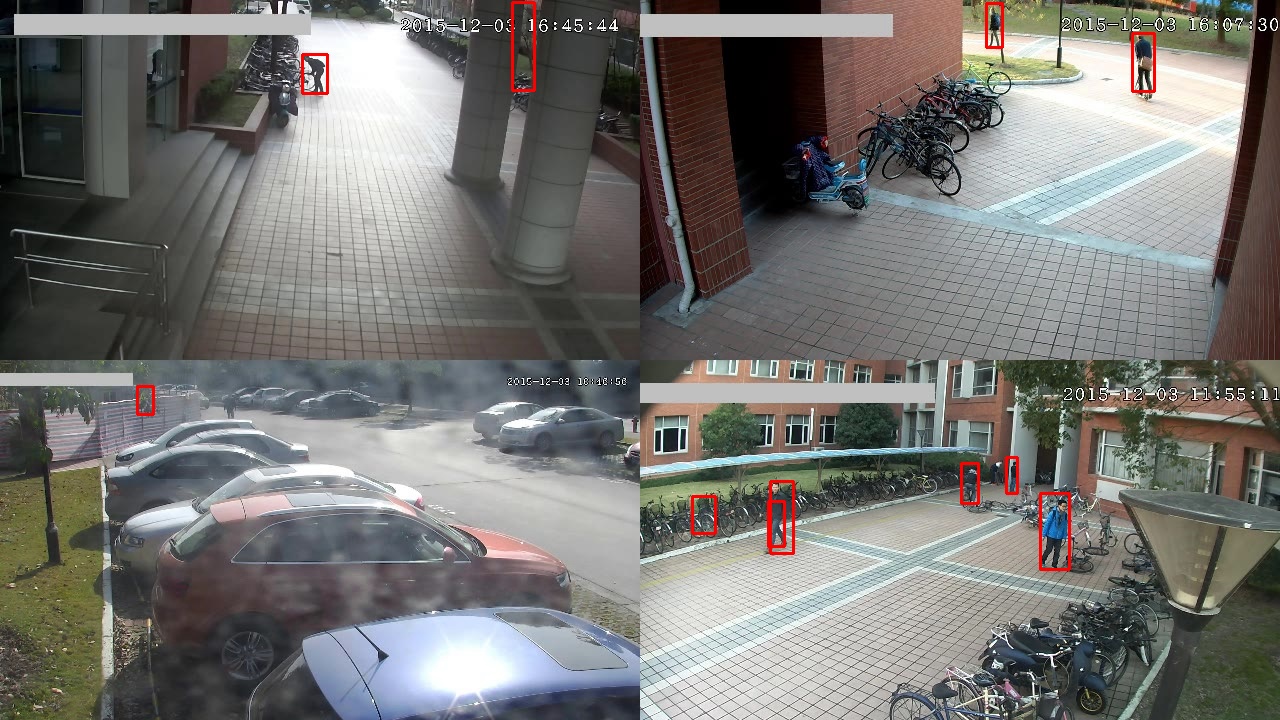}
    \label{fig:figure9b}}\\
  \subfloat[]{\includegraphics[width=4cm,height=2.8cm]{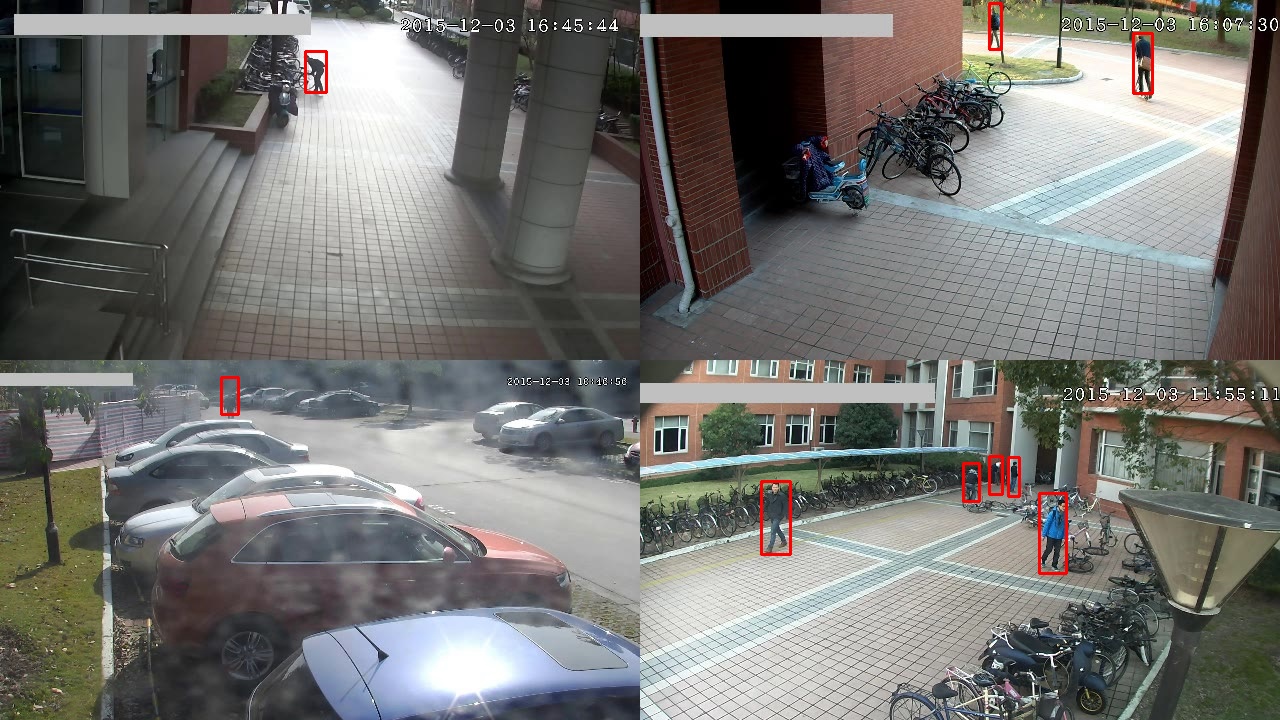}
    \label{fig:figure9c}}
  \subfloat[]{\includegraphics[width=4cm,height=2.8cm]{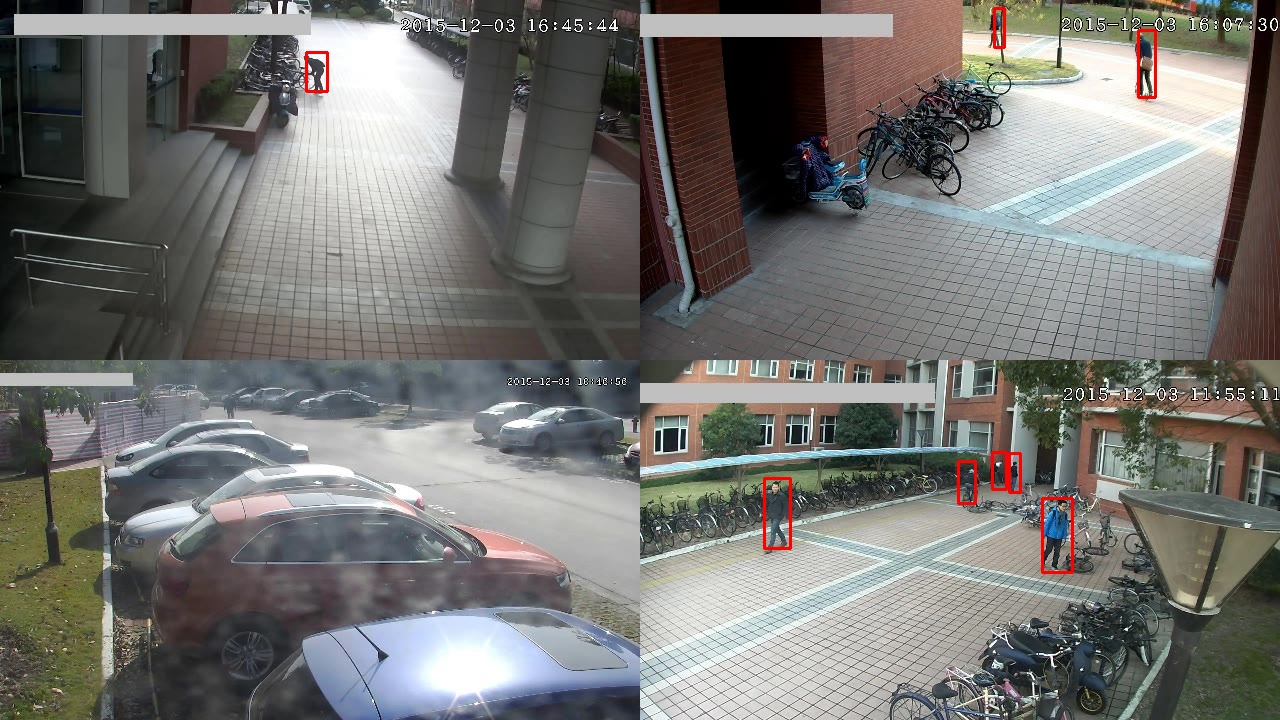}
    \label{fig:figure9d}}
  \caption{   Examples of detection results on BEST dataset.
  }\label{fig:figure9}
\end{figure}

 From Table~\ref{tab:Table1}-~\ref{table3} and Figs.\ref{fig:figure6}-~\ref{fig:figure9}, we can observe that:

 $(1)$The DS method has poor performance due to the loss of visual details for small objects. For example, we can see from Fig.~\ref{fig:figure8a} that the DS method misses many small objects.

 $(2)$The DIV method can effectively improve the detection accuracy. However, it still has two limitations: a) The computation complexity of the DIV method is high since it needs to input a large number of sub-frames into a detector (cf. the last column in Tables ~\ref{tab:Table1}-~\ref{tab:Table3}). b) Since directly dividing frames may separate one object into different sub-frames, this also results in false or repeated detection (cf. the person circled by yellow in Fig.~\ref{fig:figure7b}).

 $(3)$Compared with the DS and DIV methods, our approach (Our-S and Our-L) has obvious advantages: a) Since our approach composites sub-frames adaptively, it can effectively avoid the problem of dividing an object into different sub-frames. b) Since the adaptive patch composition process reduces the number of sub-frames, the detection speed is significantly improved from the DIV method (cf. the last column in Tables ~\ref{tab:Table1}-~\ref{tab:Table3}). c) Since our approach properly maintains the potential objects' visual information in original resolutions, the recognition accuracy is also significantly improved from the direct down-sampling method (DS).

 $(4)$Comparing Our-S method with Our-L method, since Our-L method increases the size of sub-frames, the number of sub-frames that are input into detectors are further reduced. This can obtain further improved detection speed with slightly reduced accuracy due to the down-sampling of these large sub-frames into a standard $300\cdot300$ input size of detectors.

 Moreover, Table ~\ref{table4} also shows the running time of each component in Our-L approach. We can see that the overall complexity of our approach is low. Specifically, the running time of patch extraction and patch composition components is even less than 8 ms, which is able to guarantee real-time processing.

\section{CONCLUSION\label{section:CONCLUSION}}

 In this paper, a new approach is proposed to boost both the accuracy and speed for object detection.The proposed approach first extracts patches in a video frame which are potential to include objects-of-interest, then adaptively composes the extracted patches into an optimal number of sub-frames for object detection. In this way, we are able to maintain the resolution of the original frame during object detection to guarantee the accuracy, while minimizing the number of input frames to boost the speed. Experimental results demonstrate the effectiveness of the proposed approach.

{\small
\bibliographystyle{ieee}
\bibliography{egbib}
}
\end{document}